\documentclass{article}

\usepackage{fullpage}
\usepackage{authblk}

\usepackage[T1]{fontenc}
\usepackage{graphicx}
\usepackage{float}
\usepackage[toc]{appendix}

\usepackage{booktabs}
\usepackage{amsmath,amssymb,amsthm}
\usepackage[svgnames]{xcolor}
\usepackage[colorlinks=true,bookmarksnumbered,citecolor=DarkGreen,linkcolor=DarkRed]{hyperref}
\usepackage[nameinlink]{cleveref}

\DeclareMathOperator{\ch}{\mathsf{ch}}
\DeclareMathOperator{\num}{\mathsf{num}}
\DeclareMathOperator{\den}{\mathsf{den}}
\newcommand{\eqdef}{\overset{\mathrm{\scriptscriptstyle def}}{=}}
\newcommand{\ZZ}{\mathbb{Z}}
\newcommand{\JRSZ}{\mathsf{JRSZ}}

\newcommand{\SQTPQ}{\mathsf{SQ2PQ}}

\usepackage{xkeyval}
\usepackage{xargs}

\newtheorem{example}{Example}

\begin{document}
\title{Fast Private Parameter Learning and Inference for Sum-Product Networks}
\author{Ernst Althaus\thanks{\href{mailto:ernst.althaus@uni-mainz.de}{ernst.althaus@uni-mainz.de}}
, 
Mohammad Sadeq Dousti\thanks{\href{mailto:modousti@uni-mainz.de}{modousti@uni-mainz.de}}
~and 
Stefan Kramer\thanks{\href{mailto:kramer@informatik.uni-mainz.de}{kramer@informatik.uni-mainz.de}}
~and 
Nick Johannes Peter Rassau\thanks{\href{mailto:nrassau@students.uni-mainz.de}{nrassau@students.uni-mainz.de}}
}
\affil{Institute of Computer Science, Johannes Gutenberg University Mainz, Germany}
\maketitle

\begin{abstract}
A sum-product network (SPN) is a graphical model that allows several types of inferences to be drawn efficiently.
There are two types of learning for SPNs: Learning the architecture of the model, and learning the parameters.
In this paper, we tackle the second problem: We show how to learn the weights for the sum nodes, assuming the architecture is fixed, and the data is horizontally partitioned between multiple parties. The computations will preserve the privacy of each participant. Furthermore, we will use secret sharing instead of (homomorphic) encryption, which allows fast computations and requires little computational resources. 
To this end, we use a novel integer division to compute approximate real divisions.
We also show how simple and private inferences can be performed using the learned SPN. \\[1em]

\end{abstract}

\section{Introduction}
Privacy is the ability of an individual over who stores and processes their data. It is an ever increasing concern, and recent laws such as EU's General Data Protection Regulation (GDPR) or California Consumer Privacy Act (CCPA) seek to protect individual's privacy. Prior to that, there were (and still are) laws protecting privacy in high areas of concern, such as Health Insurance Portability and Accountability Act (HIPPA), whose privacy rule protected medical records and other personal health information. These laws have strict control over how individual's data should be stored, communicated, and processed.

Privacy is often at odds with machine learning techniques, where gathering a vast amount of high-quality data is the key to success. For instance, several hospitals cannot simply share their medical records and then run a machine learning algorithm to construct some type of diagnosis model. To this end, privacy-preserving machine learning is invented, which guarantees privacy while achieving promising models with the same (or comparable) accuracy.

The problem with the above approach is that the resulting protocols are often overly complicated, which is due to the use of cryptographic constructs such as fully homomorphic encryption or oblivious transfer (see the next section). Furthermore, the performance penalty is often considerable, and sometimes the protocol is rather inefficient or even impractical.

We observed, however, that by implementing a private division protocol, some machine learning tasks can be performed privately. The division protocol can be informally explained as follows: We have $n$ parties, and party $i$ ($1\le i\le n$) has two inputs: $a_i$ and $b_i$. The protocol aims to compute the division $\lfloor \frac{a_1 + \cdots + a_n}{b_1 + \cdots + b_n} \rceil$. As an example, we show how training a sum-product network (SPN) privately and performing private inferences on it is possible. In the last section, we also discuss how private $k$-means clustering can be done.

We picked SPNs due to their operation simplicity: A SPN is a graphical model that represents a probability distribution. It is a rooted directed acyclic graph (DAG), consisting of two types of nodes: Sum nodes, and product nodes. The nodes are arranged in alternating layers. The sum nodes compute a weighted sum over the values of their children, where the weights are defined by the edges connecting a sum node to its children. The product nodes simply compute the multiplication of the values of their children.

To state the research problem rather formally, let $D$ be a dataset that is \emph{horizontally} shared between $N$ participants: $D_1, \ldots, D_n$. In other words, each participant $i$ owns a subset $D_i$ of rows in $D$ with all of the corresponding features. The participants want to jointly learn the parameters for an SPN, whose architecture is initially agreed. That is, given the architecture of the SPN, the participants want to \emph{privately} learn the weights on the outgoing edges from the SPN sum nodes (see \Cref{sec:spn}). At the end of the protocol, each party learns a secret share (\Cref{sec:ss}) of the weights on the edges, not the actual value of the weight. This increases the promise of privacy. We further show how these shares can be used to perform private inferences on the SPN. The learning protocol shall have the same result as if the whole dataset was available centrally, and the learning was done by a single party. Furthermore, at the end of the protocol, no party $i$ shall learn anything about the private inputs of other parties (i.e., any knowledge about $D_j$ for $j \ne i$).

Let us summarize the novelties in this paper:
\begin{enumerate}
\item We present a novel private division protocol that can be performed efficiently using modular additions and multiplications;
\item We show how our protocol can be used to privately learn the parameters of an SPN, and perform private inference on it. To the best of our knowledge, the only previous work in this area is CryptoSPN \cite{treiber2020cryptospn}. Compared to this work,
\begin{enumerate}
\item CryptoSPN only performs private inference. That is, it does not allow for private learning of the model.
\item CryptoSPN models the inference as a Boolean circuit, and proceeds bit-by-bit. It uses a rather complex cryptographic primitive called oblivious transfer. Our protocol, however, works directly with numbers and only uses modular additions and multiplications. 
\item CryptoSPN is limited to one client and one server (two parties), while ours is a multiparty protocol.
\item CryptoSPN is outperformed by our protocol, as shown in the experimental results.
\end{enumerate}
\item We show that our private division protocol can be used for in other machine learning tasks, such as for private $k$-means clustering \cite{jha2005privacy}. 
\end{enumerate}

The rest of this paper is as follows: \Cref{sec:prelim} explains the basic notation and the background required for the rest of the paper.
\Cref{sec:main} discusses our novel division protocol, and shows how it can be used to privately learn an SPN model.
\Cref{sec:infer} shows how private inference can be performed over a privately learned SPN.
\Cref{sec:exp} presents the experimental results.
\Cref{sec:conc} discusses the application of our novel division protocol to private $k$-means clustering, and concludes the paper.

\section{Preliminaries}	\label{sec:prelim}

\subsection{Notations}
We use $\ZZ$ to denote the set of integers. For any prime number $p$, let $\ZZ_p$ denote the (additive) group over $\{0,\ldots,p-1\}$, where the operation is addition modulo $p$. %The corresponding finite field is denoted as $\ZZ^*_p$, where the operation are addition and multiplication modulo $p$, and the underlying set is $\ZZ_p \setminus \{0\}$.

\subsection{Secure Multiparty Computations}
Assume there are $n$ parties with private inputs $x_1,\ldots,x_n$, respectively, who are interested in computing a functionality $f$, such that $(y_1,\ldots,y_n) = f(x_1,\ldots,x_n)$. That is, at the end of the computation, party $i$, $1\le i\le n$ receives $y_i$. Informally, this computation is called secure if at the end of the protocol, each party $i$ learns nothing beyond its own private input and output $(x_i, y_i)$. Rather more formally, for each party $i$, there should be a probabilistic polynomial-time machine $S$ that, on input $(x_i, y_i)$, \emph{simulates} the view%
\footnote{The \emph{view} is the set of messages received by a party during the protocol.}
of party $i$ in the protocol. In other words, each party could have computed the received messages by himself, given he knew his private input and output, and as such, no knowledge beyond that is communicated during the protocol (a formal definition can be found in \cite{goldreich2004foundations}). As such, preserving privacy is one of the most important properties of secure multiparty computations.

Secure multiparty protocols are often designed in a setting where parties are \emph{honest but curious}: Each party follows the protocol exactly as described, but might process the received messages in order to gain knowledge into other parties' private inputs. For instance, if the protocol specification a party to send a random integer, the party will do so. In a \emph{malicious} setting, the parties may arbitrarily deviate from the protocol specification. In the above example, the party may send a crafted non-random integer, in order to gain knowledge in later stages of the protocol. Protocols designed in an honest-but-curious setting can be compiled into protocols secure in the malicious setting, by adding verification at each step \cite{goldreich1987play,chor1985verifiable}. This is why our protocol assumes that all parties are honest but curious.

There are two generic approaches to securely compute \emph{any} functionality: Yao's garbled circuits (GC) protocol \cite{yao1982protocols} and Goldreich--Micali--Wigderson (GMW) protocol \cite{goldreich1987play}. Both work by modeling the functionality as a Boolean circuit, and both use a cryptographic primitive called Oblivious Transfer (OT). Informally, in a $\binom{k}{1}$-OT protocol, there is a sender who has $k$ private inputs $s_1,\ldots,s_k$, and a receiver who has one input $r \in \{1,\ldots,k\}$. At the end of the protocol, the receiver should learn $s_r$, while no knowledge about other inputs of the sender should be communicated. On the other hand, the sender learns nothing about $r$. Each OT is computationally as costly as modular exponentiation. Both Yao's GC and GMW protocols make use of many OTs (proportional to the number of gates in the Boolean circuit), although there are techniques that reduce the amortized cost \cite{asharov2015more}. It is also noteworthy that Yao's GC is essentially a two-party protocol, and recent attempts at making it a multiparty protocol are very complex even for the simple functionalities like testing equality \cite{ben2018multiparty}.

For arithmetic circuits, generic solutions are offered through a primitive called (fully) homomorphic encryption, which is introduced next. However, as we will point out, this primitive is still rather inefficient. In general, it is best to design a custom protocol rather than sticking to the generic solutions, whenever efficiency matters. In this paper, we will take this approach (custom protocol).

\subsubsection{Homomorphic Encryption}
Homomorphic encryption is an encryption function $E$, such that for any valid public key $pk$, we have $E_{pk} \colon (G,\odot) \to (H,\oplus)$, where $G$ is a group with operation $\odot$, and $H$ is a group with operation $\oplus$. The encryption function $E$ has the additional property of homomorphism: For any valid public key $pk$, and any two messages $m_1,m_2 \in G$, we have:
\begin{equation}
E_{pk}(m_1 \odot m_2) = E_{pk}(m_1) \oplus E_{pk}(m_2) \,. 
\end{equation}
This property is useful since instead of performing $\odot$ on messages, one can perform $\oplus$ on ciphertexts, and later decrypt the result using the secret key $sk$ corresponding to the public key $pk$.

It turns out that homomorphism can be exploited to achieve privacy-preserving computations. As an example, suppose both $\odot$ and $\oplus$ are simple additions, and we want to hold an e-voting between to candidates, where voters vote 0 and 1 for the first and second candidate, respectively. A \emph{voting authority} announces its public key $pk$, and voters encrypt their votes using $E_{pk}(\cdot)$. They privately submit their encrypted votes $E_{pk}(v_1),\ldots,E_{pk}(v_n)$ to a \emph{counting authority}, who does not know the secret key $sk$. The counting authority simply computes the sum of the ciphertexts $\sum_{i=1}^n E_{pk}(v_i)$, which by the homomorphism property equals $E_{pk}(\sum_{i=1}^n v_i)$. The result is submitted to the voting authority, who uses the secret key to decrypt it and announce the result $\sum_{i=1}^n v_i$.

While this shows the power of homomorphism, there is a certain limitation: Homomorphic encryption is limited to one group operation, while in many real-world applications, we need field operations (i.e., two operations like addition and multiplications). For instance, many interesting functionalities like exponentiation and logarithm can be expressed by a Taylor series over a field. If we could generalize homomorphic encryption to fields, such interesting functionalities could be computed privately.

The problem of constructing ``fully'' homomorphic encryption functions was posed as an open problem by \cite{rivest1978data}, and solved in the affirmative by \cite{gentry2009fully}. However, the initial constructions were quite slow: Computing a simple bit operation would take half an hour. There has been much improvement in the efficiency of fully homomorphic encryption constructs, yet they are still relatively slow. The interested reader is referred to \cite{Hal17} for a detailed survey of the work in this area.

\subsubsection{Secret Sharing}  \label{sec:ss}
Another approach to privacy-preserving computation is secret sharing: Here, each party splits its private input to several shares, and distributes these shares between other parties. It should be guaranteed that a share does not reveal any information about the private input of that party. However, by putting together all shares, one should be able to reconstruct the original private input.%
\footnote{In some schemes, having $k$-out-of-$n$ shares would allow the parties to reconstruct the original value, where $n$ is the number of parties, and $k \le n$ is a parameter of the scheme. Such schemes are useful if some parties refrain from taking part in the protocol in the later stages.}

In a privacy-preserving protocol based on secret sharing, the parties would locally compute the results over their shares, and occasionally exchange shares of the intermediate results. That is, at no stage are the intermediate results themselves revealed. As stated earlier, shares of any input or intermediate result should reveal no information about them. At the final stage, the shares are put together to reconstruct the final result.

Additive secret sharing over $\ZZ_p$ is a simple scheme. Here, shares of a number $x$ are $n$ values $x_1,\ldots,x_n \in \ZZ_p$ such that $\sum_{i=1}^n x_i = x \pmod p$. Furthermore, each share (except the last one) must be picked uniformly at random from $\ZZ_p$, to guarantee that they do not reveal any information about $x$. Later, we will use a protocol called \emph{joint random sharing of zero} over $\ZZ_p$, denoted $\JRSZ(\ZZ_p)$. In this protocol, a third party creates $n$ additive shares for $0$, and sends the share of each party to him. It is possible to trade a third party by some overhead \cite{catalano2005efficient}.

Another secret-sharing scheme is due to \cite{shamir1979share}, which uses polynomial secret sharing over $\ZZ_p$. To have a $k$-out-of-$n$ secret sharing scheme, a (random) polynomial of degree $k-1$ is constructed: $p(x) = c_0 + c_1x + \cdots + c_{k-1}x^{k-1} \pmod p$, where $c_0$ is the secret, and $c_1,\ldots, c_{k_1}$ are picked uniformly at random from $\ZZ_p$. The share of Party $i$, $1 \le i \le n$ equals $p(i)$. To reconstruct the secret, the parties must reconstruct the polynomial. This can be done easily as $k$ points on a polynomial of degree $k-1$ would completely determine it. Mathematically speaking, this can be done using the Lagrange interpolation formula. In the rest, we assume that $k=n$; that is, all parties should get their shares together in order to reconstruct the secret.

Shamir's secret sharing is additive: Let $S,S' \in \ZZ_p$ be two secrets, and $\{s_i\}_{i=1}^n$ and $\{s'_i\}_{i=1}^n$ be their secret shares, respectively. Then $\{s_i + s'_i\}_{i=1}^n$ are shares of $S+S'$ (note that additions and multiplications are modulo $p$). Another interesting property that we will use in this paper is that Shamir shares and additive shares over $\ZZ_p$ can be converted to one another. Specifically, we use the $\SQTPQ$ protocol of \cite{algesheimer2002efficient}, which converts additive shares over $\ZZ_p$ to polynomial (i.e., Shamir) shares.

\subsection{Sum-Product Networks (SPNs)}    \label{sec:spn}

\begin{figure}[t]
\centering
\includegraphics[width=0.5\columnwidth]{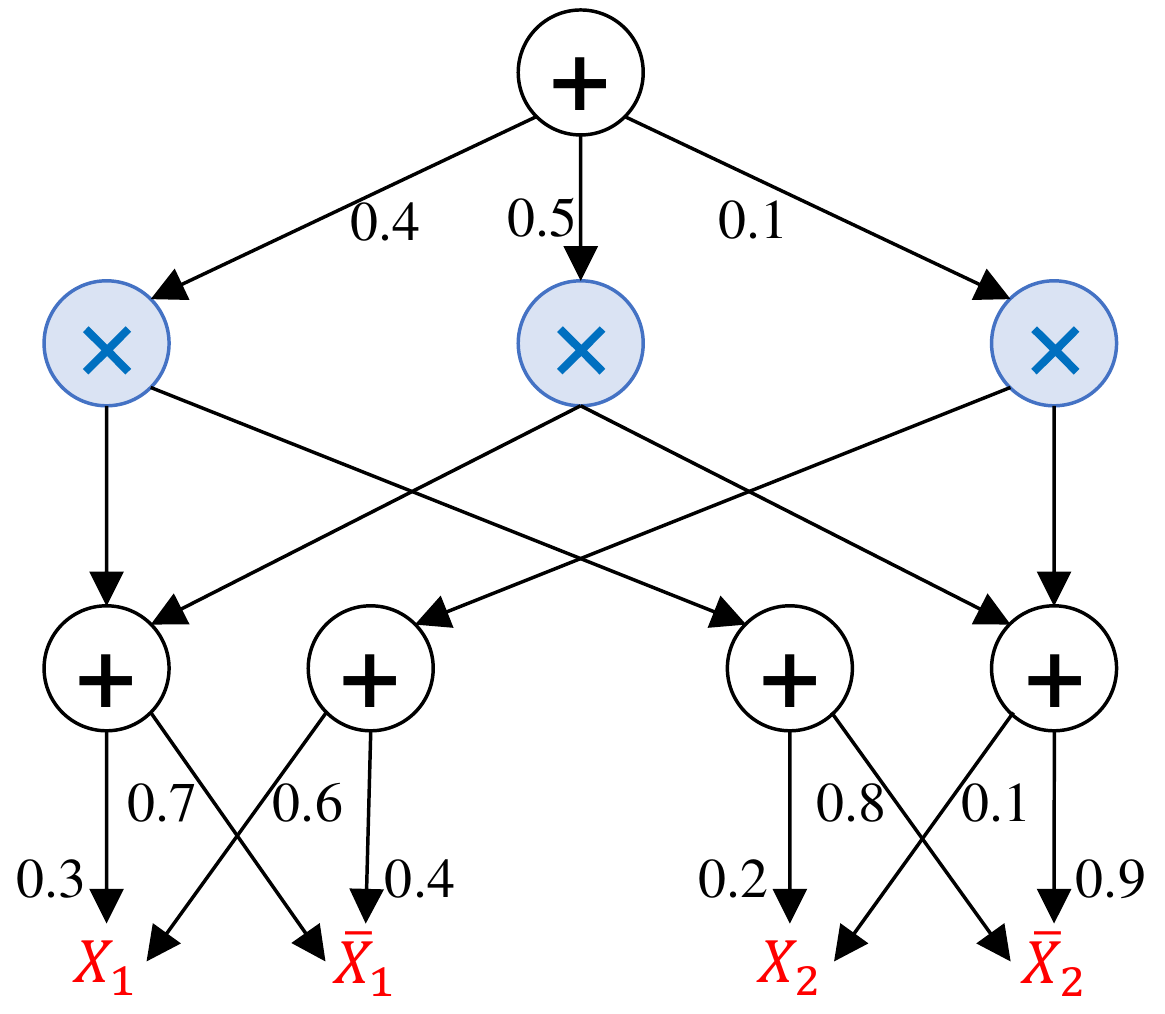}
\caption{A small SPN network with two indicator random variables $X_1$ and $X_2$.}
\label{fig:spn}
\end{figure}

A Sum-Product Network (SPN) is a rooted acyclic graph where the leaves are indicator%
\footnote{Using any univariate (not necessarily indicator) random variables is also considered in the literature.}
random variables and their complements, connected to sum nodes. The sum nodes are connected to product nodes, and alternating layers of sum and product nodes ensue. The outgoing edges from sum nodes are weighted, and the sum of weights on all outgoing edges from a sum node is 1. The {\em scope} of a leaf node with a variable $V$ is $\{V\}$, the scope of a non-leaf node is the union of the scopes of its children. If all sum nodes have children all sharing the same scope, i.e., they are {\em complete}, and all product nodes have children having disjoint scopes, i.e., they are {\em decomposable}, then the network represents a unique probability function.

As an example, consider \Cref{fig:spn}: The sum node in the bottom layer and to the far left computes $0.3 X_1 + 0.7 \bar{X}_1$. The network as a whole computes $S$ as denoted below:
\begin{align*}
S_1 &= 0.3 X_1 + 0.7 \bar{X}_1 &
S_2 &= 0.6 X_1 + 0.4 \bar{X}_1 \\
S_3 &= 0.2 X_2 + 0.8 \bar{X}_2 &
S_4 &= 0.1 X_2 + 0.9 \bar{X}_2 \\
P_1 &= S1 \times S_3 &
P_2 &= S1 \times S_4 \\
P_3 &= S2 \times S_4 &
S &= 0.4P_1 + 0.5P_2 + 0.1P_3 
\end{align*}

There are various graphical models expressing the conditional dependence structure between random variables, such as the Bayesian networks. However, SPNs have the advantage that several types of inference can be performed in time that is a polynomial function of the number of edges in the graph. These efficient inferences include computing marginal and posterior probabilities, most probable explanation (MPE), approximate maximum a-posteriori (MAP), and approximate MAX. For a good and detailed introduction to SPNs and these types of inferences, the reader is referred to \cite{sanchez2021sum}.

To the best of our knowledge, there is only one previous work regarding privacy-preserving SPNs, called \textsc{CryptoSPN} \cite{treiber2020cryptospn}. In that work, the authors use Yao's garbled circuit for a two-party protocol between a client and a server. The server has the SPN model, and the client has some data. At the end of the protocol, the client received the inferences on its data based on the SPN model on the server side. In this work, we take the next step: We allow the joint private learning of the SPN model (i.e., model parameters), and then allow inferences to be drawn. We assume the structure of the SPN is already fixed.

\section{Privacy-Preserving Parameter Learning for SPNs}	\label{sec:main}
\subsection{Parameter Learning}
Let us denote the nodes of SPN graph by $n_1, n_2, \ldots$. Whenever there is an edge $n_i \to n_j$, we call $n_j$ the child of $n_i$. The set of children of $n_i$ is denoted by $\ch(i)$. If $n_i$ is a sum node, the outgoing edge to any of its children $n_j$ has a weight $w_{ij}$. We assume the weights are non-zero, and the sum of all outgoing weights from a sum node is 1. We further assume two structural properties of the SPN: (1) \textbf{Completeness:} All sum nodes have children whose scopes are the same (i.e., defined over the same model variables), and (2) \textbf{Decomposability:} All product nodes have children whose scopes are disjoint. Furthermore, we suppose the SPNs are (3)~\textbf{Selective} \cite{peharz2014learning}: For each sum node, at most one child makes a positive contribution. This constraint allows for finding the optimal parameters in closed form.

For each sum node $n_i$ and any $n_j \in \ch(i)$, define $n_{ij}$ as the number of instances in the dataset where $n_j$ makes a positive contribution to $n_i$. It is shown that the weight $\hat{w}_{ij}$ that maximizes the likelihood can be obtained in closed form as follows; see Equation (24) of \cite{sanchez2021sum}:
\begin{equation}	\label{eq:wij}
\hat{w}_{ij} = \frac{n_{ij}}{\sum_{j' \in \ch(i)} n_{ij'}}
\end{equation}

Consider an honest-but-curious setting, in which the data is horizontally partitioned between $N$ parties. The crucial observation for a multiparty computation is that each party can locally compute $n_{ij}$ locally and based on the instances in its local dataset. The global value for $n_{ij}$ (i.e., the value for the union of local datasets) is the sum of the corresponding local values.

Since the data is horizontally partitioned, the $k$th party can compute the numerator and denominator of \Cref{eq:wij} locally, which we call $\num_{ij}^{k}$ and $\den_{ij}^{k}$, respectively. Therefore, $\hat{w}_{ij}$ can be written as:

\begin{equation}	\label{eq:local}
\hat{w}_{ij} = \frac{\sum_{k=1}^N \num^k_{ij}}{\sum_{k=1}^N \den^k_{ij}}
\end{equation}
This shows that private learning the parameters of an SPN can be done by privately performing the above division.

We show three approaches to perform the division. The first one is an approximate solution, and assumes an ``almost identical'' distribution of data among the parties. This assumption is unrealistic in practice, and we included this section just for the sake of providing the reader with some numerical example. The second approach uses homomorphic encryption, and is therefore slow. It uses an existing protocol, and we only sketch it for the sake of completeness. The third protocol is the main contribution of this paper, and implements the division using a combination of additive secret sharing over $\ZZ_p$ and Shamir secret sharing.  

Notice that $0 \le \hat{w}_{ij} \le 1$ and we represent all numbers as small integers. Hence, instead computing $\hat{w}_{ij}$, we will compute an approximation of $d\cdot \hat{w}_{ij}$ for a suitable choice of a normalization factor $d$ in all approaches. We assume throughout this section that all numbers that are given into our protocols are integers in $[0, \dots, d-1]$, sometimes given in polynomial shares. We will guarantee that the results are again integer numbers in $[0, \dots, d-1]$. We assume that the prime $p$ is larger than $d$.%\textcolor{red}{Does this suffice in all approaches or has $p$ to be larger?}

\subsection{An Approximate Solution}
The approximate solution can be used when the dataset is large enough, and the distribution of data between the parties is almost identical.
Then, \Cref{eq:local} can be approximated as follows:
\begin{equation}	\label{eq:local-fraks}
\hat{w}_{ij} \approx \frac{1}{N} \sum_{k=1}^N \frac{\num^k_{ij}}{\den^k_{ij}}
\end{equation}
That is, the $k$th party computes the fraction $f^k_{ij} \eqdef \frac{\num^k_{ij}}{\den^k_{ij}}$ locally, and $w_{ij}$ is approximated as the average of the local fractions.

To benefit from the later evaluation protocols that use secret sharing, each party must hold a uniformly distributed share of $\hat{w}_{ij}$. This can be done by choosing a suitable large prime number $p$, a normalization factor $d$, and executing the following protocol:
\begin{enumerate}
\item \textbf{Preprocessing:} This part is independent of each party's private input, and can be done in a preprocessing step. The $N$ parties take part in a \emph{joint random sharing of zero} over $\ZZ_p$, denoted $\JRSZ(\ZZ_p)$. At the end, party $k$ obtains $r^k_{ij} \in \ZZ_p$, such that $\sum_{k=1}^N r^k_{ij} = 0 \pmod p$.
\item Party $k$ computes $F^k_{ij} \eqdef \lfloor \frac{d \cdot f^k_{ij}}{N} \rceil$. This should be less than $p$ assuming $p$ is large enough.
\item Party $k$ computes $\hat{F}^k_{ij} \eqdef F^k_{ij} + r^k_{ij} \pmod p$ as an additive, secret share of $\hat{w}_ij$.
\end{enumerate}

\begin{example}
Consider three parties, $p=2^{20}+7$, and $d=1000$. For simplicity, we drop the $ij$ subscripts. 
Let $(r^1, r^2, r^3) = (752508, 776879, 567779)$ be the result of $\JRSZ(\ZZ_p)$, and assume the numerators and denominators are as follows:
\begin{align}
(\num^1, \num^2, \num^3) &= (71, 209, 320) \\
(\den^1, \den^2, \den^3) &= (256, 786, 1127)
\end{align}
Then $\hat{w} = \frac{71 + 209 + 320}{256 + 786 + 1127} = 0.277$, and the approximation is
$\frac{1}{3}(\frac{71}{256} + \frac{209}{786} + \frac{320}{1127}) = 0.276$, which is pretty close.

The parties compute their fractions as $(F^1, F^2, F^3) = (92, 89, 95)$. Notice that the sum of $F^k$ divided by $d$ gives a value equal (or very close) to the approximation $0.276$.

The final shares are $(\hat{F}^1, \hat{F}^2, \hat{F}^3) = (752600, 776968, 567874)$. It can be verified that $\sum_k \hat{F}^k = 276 \pmod N$, which is $0.276$ when divided by $d$.
\end{example}

\subsection{An Exact Solution Based on Homomorphic Encryption (Sketch)}
Let $(pk,sk)$ be the public and secret keys of a fully-homomorphic encryption scheme, generated by a third party. This third party publishes $pk$ to all other parties.

Party $i$, $1\le i \le N$ computes $E_{pk}(d \cdot \num_i)$ and $E_{pk}(\den_i)$, where $d$ is an integer as in the previous subsection. All parties then send these values to Party 1, who uses the additivity of of encryption to compute
\begin{align*}
\sum_{i=1}^N E_{pk}(d \cdot \num_i) &= E_{pk}(\sum_{i=1}^N d \cdot \num_i) = E_{pk}(d \cdot \num) \cr
\sum_{i=1}^N E_{pk}(\den_i) &= E_{pk}(\sum_{i=1}^N \den_i) = E_{pk}(\den)
\end{align*}

To compute the weights, one can apply the division method of \cite{cetin2015arithmetic}.

\subsection{An Exact Solution Based on Secret Sharing}
In this section, we show how to compute an exact division adopting the method proposed by \cite{algesheimer2002efficient} to compute an approximate inverse of a shared number scaled with a public normalization factor. That is, given shares of $b$, we compute shares of a close approximation of $d/b$. (Recall from the previous sections that $d$ is a number that adjusts the precision of division.) Given this protocol, the parties owning shares of $a$ can perform a secure multiplication to compute shares of $a \times d/b$. Later, when if they want shares of $a / b$, they can simply perform a secure truncate to remove $d$. The protocols for secure multiplication and secure truncate are explained in \cite{algesheimer2002efficient}.

The difference of approximate inverse to cited work is that we avoid converting between different representations of the numbers, as we will only use polynomial shares, and thus our approach is more efficient. More importantly, we do not assume to know an initial guess $u$ with $d/2b \le u \le d/b$ in the beginning. This is of utmost importance to the correctness of our protocol, since the parties who have shares $b$ do not know an exact upper or lower bound on it.

When describing the protocol, we denote a number $x$ which is only known in polynomial shares over $\ZZ_p$ with $[x]_p$. As in \cite{algesheimer2002efficient}, we will basically use the Newton-method to compute an approximate root of the function $f$ with $f(x)=1/x - b/d$ which is an approximation of $[u]_p \approx d/b$. Recall that a single step of the Newton-method given a current approximation $u$ is $u \leftarrow u-f(x)/f'(x)=u+(1/u-b/d)/(1/u^2)=u(2-ub/d)$.
To perform a single step of the Newton-iteration given shares $[u]_p$ of a current approximation $u$ for $d/b$, we need to have a secret sharing method to compute additions and multiplications of shares and a division of a shared number by a public number. Notice that the division causes a rounding error, if we stick to integers. As addition and multiplication is well known, we only give a method for the (approximate) division by a public number.

\cite{algesheimer2002efficient} showed that we have a quadratic convergence if we start with an approximate $u$ with $u/2 \le d/b \le u$ and increase the precision in each round by a certain factor $e$. More precisely, they replace an interation by $u \leftarrow u(2-ud/(de))$ and showed that for any $t > \log(5+\ln(k+1))$, after $\log(e)$ steps, we have an approximation of $de/b$ with a relative error of at most $16(k+1)/e$, if the difference of the computed division by a public number $d$ and the real number is at most $k+1$. %\textcolor{red}{TODO verify!}

As we do not have a number $u$ with $d/2b \le u \le d/b$, we start with $u=1$; i.e.,~an underestimation of $u$ and how that after $\lceil \log(d) \rceil$ iterations, we have $d/2b \le u \le d/b$. This can be seen as follows. Let $u_i$ be the approximation of $d/b$ after the $i$-th iteration and $f_i=d/bu_i$. As $u_{i+1}=u_i(2-u_ib/d)=u_i(2-1/f_i)$, we have $f_{i+1}=f_i f_i/(2f_i-1)$. Hence, if $f_i \le 2^n+1$, we have $f_{i+1} \le 2^{n-1}+1$. As $f_0 \le d+1$, we have $f_{\lceil \log(d) \rceil} \le 2$.
%\textcolor{red}{TODO verify!}
Hence, we perform the Newton-method for additionally $\lceil \log(d) \rceil$ steps. %\textcolor{red}{TODO: should we keep to following:} Notice furthermore that we will not detect a division by zero but return approximately $d$ as a result. As for parameter learning, the nominator is at most the denominator, we will multiply this result by zero anyway, resulting in $w_{ij}=0$ if the nominator and denominator are zero.\textcolor{red}{What is usually done for SPNs in this case?}

We still have to show how to compute a number close to $[u]_p/d$ for a shared value $u$ and a public value $d$. \cite{algesheimer2002efficient} showed a method that is based on first computing the so-called \emph{integer shares of $u$}, which is only secure up to a security parameter $\rho$. We propose an alternative method that is more efficient, but also secure only up to a security parameter $\rho$. The protocol involves two special parties, called Alice and Bob in the following, and works as follows:
\begin{itemize}
    \item Alice generates a number $r$ from $[0, \dots, 2^\rho-1]$ uniformly at random, computes $q= r \mod d$ and distributes polynomial shares $[r]_p$ and $[q]_p$
    \item The parties compute $[z]_p=[u]_p + [r]_p$ and reveal the number to Bob, i.e.~Bob now knows $z$.
    \item Bob computes $w=z \mod d$ and distributes polynomial shares $[w]_p$
    \item The parties compute $[z]_p=[u]_p-[q]_p+[w]_p$ and obtain the result by $[z_p] \cdot d^{-1}$, where $d^{-1}$ is the multiplicative inverse in $\ZZ_p$.
\end{itemize}

The correctness can be shown as follows. We first notice that $z \mod d = u \mod d - q \mod d + w \mod d=u \mod d + r \mod d - (r+u) \mod d =0$. Furthermore $u-d \le z \le u+d$, as $\mod$ gives a number between $0$ and $d-1$. As $u \mod d=0$, the multiplication with $d^{-1}$ in $\ZZ_p$ is equal to the division with $d$ in $\ZZ$. Hence, we end up with a number in $[u/d-1, u/d+1]$.

The security can be seen as follows. The only number that is revealed in the protocol is $z=r+u$, which will be known only to Bob at the end. As Bob knows nothing on $r$ except that $r$ is chosen uniformly at random from $\{0, \dots, 2^\rho-1 \}$, he can not obtain any information on $u$ as long as $z \in [d, 2^\rho-1]$. The probability for $z$ not lying in this interval is at most $d/2^\rho$ independently of $u$, which is negligible in $\rho$.

\section{Privacy-Preserving Inference for SPNs} \label{sec:infer}
In this section, we focus on one type of inference for SPNs: Computing marginal probabilities. Here, a set of servers share the weights of a jointly learned SPN, and a client has two configurations $\mathbf{x}$ and $\mathbf{e}$ for the leaf nodes of the SPN. The client wants to compute $\Pr(\mathbf{x} \mid \mathbf{e}) = S(\mathbf{x}\mathbf{e}) / S(\mathbf{e})$, where $S(\cdot)$ is the root value of SPN when setting the leaf nodes to the given configuration. For further details, see Section IV.A of \cite{sanchez2021sum}. The objective is that the client does not learn the network parameters (beyond what can be inferred from the inference result), and the servers do not learn the private inputs of the client. Below, we explain how $S(\cdot)$ can be computed in this manner.

For private evaluation, a client shares its private input $\mathbf{z}$ with the $N$ servers. If we are in the approximate setting, the secret sharing will be over $\ZZ_p$; otherwise, it will be over polynomials.

For sum (resp. product) nodes, the servers should simply perform a secure sum (resp. secure multiplication). However, the input to sum nodes are not that straightforward, since the weights are shared. For instance, in the shared setting over $\ZZ_p$, the servers must privately perform the value $(\sum_{k=1}^N w_{ij}^k)\cdot(\sum_{k=1}^N z_{ij}^k)$ for a given edge. This can be done with a secure multiplication of shares.
\begin{itemize}
    \item For the $\ZZ_p$ setting, the servers might use the multiplication algorithm of \cite{xiong2020efficient}.
    \item For the polynomial setting, \textsc{Sharemind} \cite{bogdanov2008sharemind} gives an improved version of Du-Atallah multiplication protocol.
\end{itemize}

\section{Experimental Results}	\label{sec:exp}
In this section, we explain the results of our experiments. Further details on the implementation can be found in \Cref{apx:impl}.

\subsection{Experiments target}
The target of the experiments is to measure the required time for private training of the network weights and the scalability as the number of parties increase. Furthermore, we will measure the number of messages sent over the network together with its size.

\subsection{Network structure}
In our Python implementation, there is a task scheduling server called the \texttt{Manager}, and multiple task executing servers called \texttt{Member}s. They are all connected to each other via the WebSocket\footnote{\url{https://websockets.readthedocs.io/en/stable/}} Framework. All members and the manager have a unique \texttt{ID} in the network, which is used for Shamir secret sharing too. Tasks to be done by the network are scheduled by the manager as \texttt{Exercise}s, which we will see later in the tables.

\subsection{Experimental settings}
For our experiments we used ``Windows Subsystem for Linux'' (WSL 2) with a specification of 32GB RAM (DDR-4), Intel i7-8700K as CPU, an internal network latency of 10~ms between each communication through the network, and an NVME Samsung SSD 970 Pro 512GB.

The parameter $n$ for the Newton iterations and the truncation is set to $16$, $t$ to $5$ and our multiplicative factor $d$ to scale real values up is set to $256$. As prime number we are using 13558774610046711780701 In the following we are focused on four datasets:\footnote{\url{https://github.com/arranger1044/DEBD} with checkout: 80a4906dcf3b3463370f904efa42c21e8295e85c} nltcs, jester, baudio and bnetflix based on \cite{datasets}.

\subsection{Results}
The structural values, such as the amount of product nodes and layers as well as parameters can be found in the \Cref{table:Dataset stats}. They are then fed into our network, and the network measurement is done twice: First with 13 \texttt{Members}, whose results are shown in \Cref{table:experimental13} and then with only five \texttt{Members}, resulting in \Cref{table:experimental5}.

\begin{table}[H]
\centering
\caption{Statistics of the used SPN structure, learned via SPFlow \cite{spflow} from the datasets}
\label{table:Dataset stats}
{\footnotesize\begin{tabular}{c c c c c c c}
\toprule
Dataset & sum & product & leaf & params & edges & layers \\ [0.5ex] 
\midrule
nltcs    & 13 & 26 & 74 & 100 & 112 & 9 \\ 
jester   & 10 & 20 & 225 & 245 & 254 & 5 \\
baudio   & 17 & 36 & 282 & 318 & 334 & 7 \\
bnetflix & 27 & 54 & 265 & 319 & 345 & 7 \\ [1ex]
\bottomrule
\end{tabular}}
\end{table}

\begin{table}[H]
\centering
\caption{Training runtime and network traffic for 13 members and one manager with latency 10~ms.}
\label{table:experimental13}
\begin{tabular}{c c c c}
\toprule
Dataset & Amount messages & size(mb) & time(s) \\ [0.5ex] 
\midrule
nltcs & 4.231.815 & 170 & 6952 \\ 
jester & 3.290.901 & 133 & 5622 \\
baudio & 5.800.005 & 233 & 9088 \\
bnetflix & 8.622.747 & 347 & 15640 \\ [1ex]
\bottomrule
\end{tabular}
\end{table}

\begin{table}[H]
\centering
\caption{Training runtime and network traffic for 5 members and one manager with latency 10~ms.}
\label{table:experimental5}
\begin{tabular}{c c c c} 
\toprule
Dataset & Amount messages & size (MB) & time (s) \\ [0.5ex] 
\midrule
nltcs & 915.273 & 36 & 2101 \\ 
jester & 711.813 & 28 & 1640 \\
baudio & 1.254.423 & 49 & 2880 \\
bnetflix & 1.864.893 & 73 & 4344 \\ [1ex]
\bottomrule
\end{tabular}
\end{table}

It is important to note that learning the model parameters is done only once during the lifetime of the system. Therefore, running times of one or two hours for a large dataset is quite justifiable.

% In the table x we see the network measurements for evaluating the network with our trained weights for 5 inputs.

\section{Discussion \& Conclusion}	\label{sec:conc}
In this paper, we showed how private learning an SPN is possible through an improved secure division protocol. We also considered the private inference, a protocol that has been solved previously using another approach (Yao's garbled circuits) by \textsc{CryptoSPN} \cite{treiber2020cryptospn}. The latter approach is inherently slower than ours, as it is suited for generic protocols. It is normally used for two-party protocols, while our approach enables multiple parties to interact. We should experimental results, confirming our protocols are justifiably fast in practice.

Our secure division protocol can also be used for other learning tasks. For instance, the private $k$-means protocol requires it as a primitive:
In \cite{jha2005privacy}, two protocol for jointly computing the following functionality are proposed:
\begin{equation}
((x_1, x_2), (y_1, y_2)) \mapsto (\frac{x_1 + y_1}{x_2 + y_2}, \frac{x_1 + y_1}{x_2 + y_2})
\end{equation}
The first protocol is based on a primitive called oblivious polynomial evaluation (OPE), and the second is based on homomorphic encryption. We effectively compute the same functionality using different techniques. Our protocols can therefore be useful in the context of privacy-preserving clustering as well.

For the future work, one can improvement the sub-protocols used. Furthermore, private learning the structure of the SPN can be considered.

\bibliographystyle{unsrt}
\bibliography{refs}

\appendix
\section{Implementation Details}    \label{apx:impl}
The operations that the network can perform are wrapped in \texttt{Exercise}s. They have an ID to determine the type of operation to be done, and a value that contains mostly arguments needed by the members to run the required operation. Thus, before some calculation takes place, the manager first builds up the order of operations to run, and then he starts scheduling the exercises.

For example, let be number $a$ be saved under the data ID \verb|data_id_a| and $b$ as 
\verb|data_id_b|, both in polynomial shares, so every 
\texttt{Member} has one share of the number. The 
\texttt{Manager} would call
\texttt{add\_exercise\_addition(data\_id\_a, data\_id\_b, data\_id\_result)} to enqueue the addition operation in the exercise queue, where the result of the addition will be stored at 
\verb|data_id_result|. When the network finishes performing all previously enqueued operations, the addition is finally scheduled to run by the network. To this end, every 
\texttt{Member} performs the \texttt{addition(data\_id\_a, data\_id\_b, data\_id\_result)} method locally, which actually executes the addition of the shared numbers. If a 
\texttt{Member} is done with its local part, it sends back the 
\texttt{ID} of the exercise together with his own network 
\texttt{ID}, thus the manager knows that the 
\texttt{Member} has finished the exercise. After receiving this ``finished'' message from all 
\texttt{Members} for this exercise, the manager will start scheduling the next exercise in the queue.

\end{document}